\documentclass[10pt,twocolumn,letterpaper]{article}

\usepackage{cvpr}
\usepackage{times}
\usepackage{epsfig}
\usepackage{graphicx}
\usepackage{amsmath}
\usepackage{amssymb}

\usepackage{algorithm, algorithmic} 
\usepackage[breaklinks=true,bookmarks=false]{hyperref}

\cvprfinalcopy 

\def\cvprPaperID{****} 

\begin{document}

\title{Unsupervised Visual Representation Learning with Increasing Object Shape Bias}

\author{Zhibo Wang\\
College of Engineering and Computer Science\\
University of Central Florida\\
{\tt\small zhibo.wang@knights.ucf.edu}
\and
Shen Yan\\
Computer Science and Engineering Department\\
Michigan State University\\
{\tt\small yanshen6@msu.edu}
\and
Xiaoyu Zhang\\
College of Engineering and Computer Science\\
University of Central Florida\\
{\tt\small x.zhang@Knights.ucf.edu}
\and
Niels Lobo\\
College of Engineering and Computer Science\\
University of Central Florida\\
{\tt\small niels@cs.ucf.edu}
}

\maketitle

\begin{abstract}
	Traditional supervised learning keeps pushing convolution neural network(CNN) achieving state-of-art performance. However, lack of large-scale annotation data always bother researchers due to the high cost of it, even ImageNet dataset is over-fitted by complex models now. The success of unsupervised learning method represented by the Bert model training in natural language processing(NLP) field shows its great potential. And it makes that unlimited training samples becomes possible and the great universal generalization ability changes NLP research direction directly. In this article, we purpose a novel unsupervised learning method based on contrastive predictive coding in computer vision field. By learning reconstruable high level representation information of object from image itself, we are able to train model with any non-annotation images and improve model's performance to reach state-of-art performance at the same level of model complexity. Beside that, the learning ability of feature representation information is common if the number of training images could be unlimited amplification, an universal large-scale pre-trained computer vision model is possible in the future.
\end{abstract}

\section{Introduction}
Deep learning model, characterized by large volume of parameters, strong domain adoption and great generalization ability, has reached state-of-art performance in every computer vision tasks and dominated this area for almost ten years long. Beyond that, new designations of model structure and massive parameters keep improving models' performance, but behind this, increasingly powerful computing power and large-scale annotation data sets are the biggest driving force. Nowadays, powerful computing power is easy to obtain, but not for high quality annotation data sets. Even now, ImageNet started to be overfitted by large scale convolution neural model, and model's performance doesn't improve much in these two years. 

The two main strategies to solve problems are: (1)  Discovering a better model architecture designation and (2) Improving model's feature extraction ability by more training data though applying extra data sources or data augmentation. For strategy 1, the mainstream research focus on minimizing the size of model but remain similar performance, the common methods are  knowledge distillation(Use pre-trained large model to teach small size model) and EfficientNet(scaling model to be best fit on specific data set). For strategy 2, by modifying the training images with computer vision to expand original training data size, so that model can get better generalization ability. Both strategies have made the models reach the best performance on Imagenet in 2019, but they don't solve the problem that lack of large-scale annotation data to train large size model, and specialized training strategy makes the model hard to overcome domain adoption problem to be an universal pre-train model. 

To overcome this challenge, semi-supervised learning which also is widely used in knowledge distillation field tries to do machine annotation by large model or web supervision to expand training data set to reach billion-scale. However, the drawback of these approaches is that these generated annotations are noisy and limited in specific categories, this also limits the model's performance improvement space. In order to avoid this situation, unsupervised learning is purposed and actively being researched on by the research community. Compared with previous research, the learning object is not decided by image's annotation in unsupervised learning strategy, training purpose is more focused on universal feature detection and extraction. The evaluation criteria for training is generated by the training image itself. 

The most common unsupervised learning strategy is to make prediction for the missing patches of contextual information in a text sentence or pixel in a image, so is also inferred as representation learning. One of the oldest strategy in this research field is generated from signal data compression called predictive coding. Inspired by this, large NLP pre-training model Bert which makes prediction by neighbor words has been proved successful in practice. So for computer vision research, since object has its own unique texture and shape characterizes, it is reasonable to assume that random pixel or image patches in the image is highly dependent on its neighbors as well on the similar shared high level latent information. And recently research has shown that this brings stronger feature extracting ability than normal supervised training result on the same data set. However, current research ignores image object's own special structure and texture information and location information. We address these key challenge by introducing new contrastive predictive method and special data augmentation.

Our first contribution is to reform contrastive predictive method's learning mechanism.  Given a image, model will be forced to pay more attention to the object's shape and profile, other than 

Our second contribution is to introduce transformer model as autoregreesion model to perform supervised train computer vision model.

Our third contribution is that we use neural transfer model as data argumentation method to regularize model's learning direction. 

Through our unsupervised training method, model shows better domain adoption ability and performance than model with regular supervised training. This also hints that an universal pre-trained large computer vision model is potentially.



\section{Related Work}

\subsection{Contrastive Predictive}

\begin{figure}
	\begin{center}
	\includegraphics[width=0.4\textwidth]{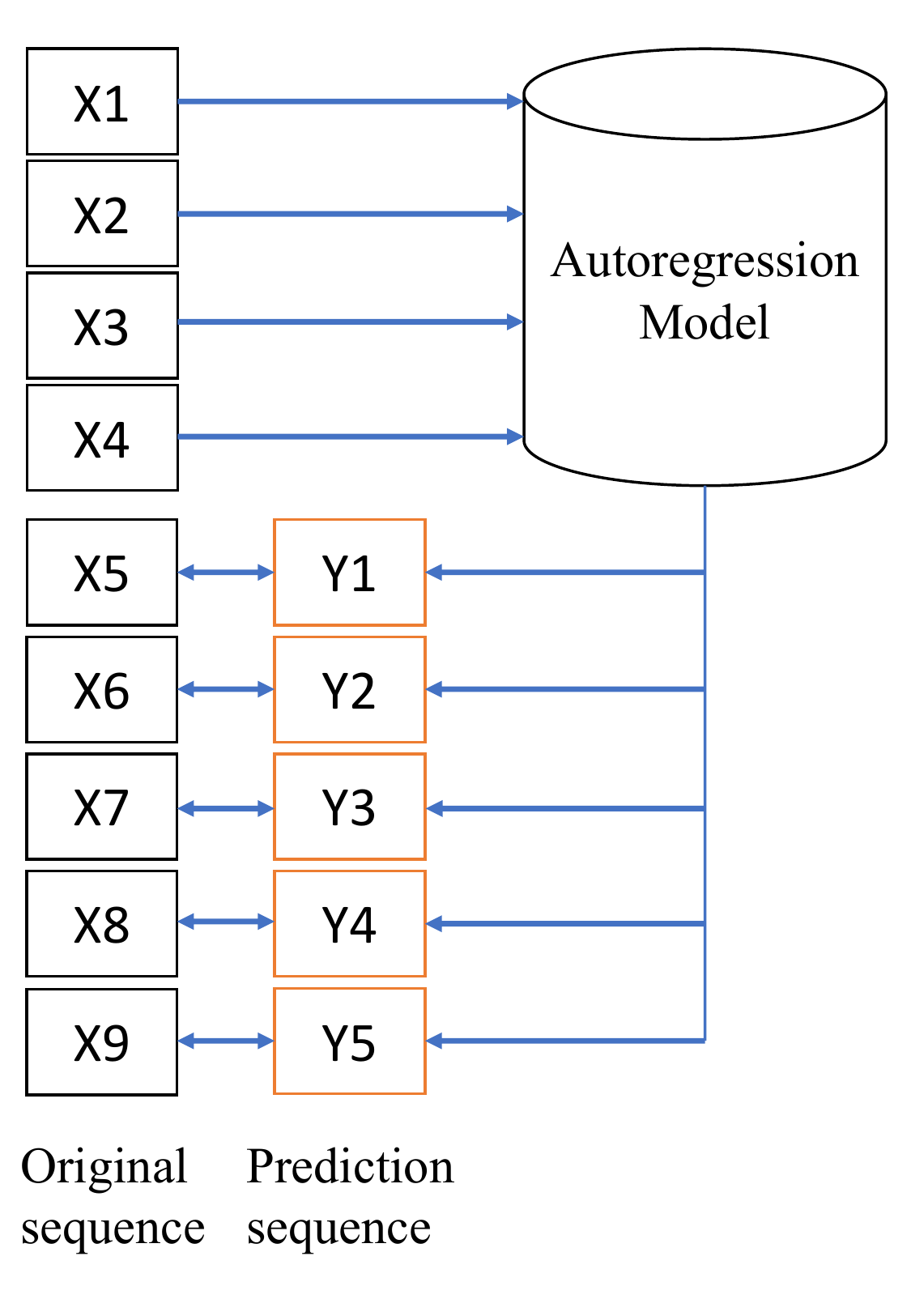}
	\end{center}
	\caption{The overview of traditional contrastive prediction coding application on a sequence data. With a sequence of known representation learning information, the second half of the sequence is masked and predicted by the autoregression model  based on the unmasked sequence part. By comparing the prediction and the known information sequence, this contracts constrastive learning which is supervised. }
	\label{fig:contrastive}
\end{figure}

Inspired by the brain neuron, constrastive predictive coding, as shown in {\bf figure \ref{fig:contrastive}}, is an unifying framework for understanding redundancy reduction and efficient coding in the nervous system.  In recent researches, it is used in pixel recurrent neural networks, video generation and universal NLP model training. With the assumption that information of data points(context or image patches or pixels) with closed locations are related and predictable by each others, the mechanism of it is to predict the future or missing information by the known information around the target. Due to the future or missing information is included or implied in the training data itself, this constructs a prefect supervised learning situation under unsupervised learning. Since a good representation of context or image should be able to reconstruct the object itself and predict its profile. In other words, a well trained model should filter useless low-level details of visual perception or context which has no help for logical inference. The mechanism of the constrastive predictive coding will regularize the model's training to meet his purpose. Recently, its related applications like GPT, Bert and previous application like word2vector all provide strong model performance and are widely used in many tasks. 

Contrastive predictive coding normally includes three parts, the first is feature representation information extraction, the next one is sequence prediction, the last is contrastive loss. 

Feature representation information extraction is normally performed by the encoder part of the computer vision model in the image processing field, and the primary object is to improve model's feature information extraction ability. 

Due to training and target data is in the sequence format, the prediction task is performed by the autoregression models. Autoregression model can be in different formats. In the recent researches, recurrent neural network(RNN) and Long Short-Term Memory(LSTM) are commonly used to compose the autoregression model to help regularize the computer vision model's training process.

Contrastive loss function is widely used in objects detection tasks and domains adoption, normally it is based on the triplet losses and using max-margin method to separate positive examples from negative. Though this loss function, the useful vision feature information can be distinguished from low-level texture features.

\subsection{Object Shape Bias Increment}

Convolutional Neural Networks (CNNs) has been successfully used in countless computer vision tasks. The most common thought about how the CNN recognizes the vision objects from its perceptions is based on the learned representation of objects' shape and texture. However, recent research proves that object's texture is much more decisive than object's shape(Object's shape's identification is also kind based on texture). This subverts our perception of the CNN's working mechanism. This also explains to some extent that why CNN lacks the generalization ability that it should have in domain adoption problems and pre-trained model is not that necessary in some cases. However, in contrast, pre-trained Bert model reached state-of-the-art in every NLP tasks at the moment it was developed. After that, recent researches prove that external training for increasing object shape learning does considerably improve model's classification performance.

\subsection{Neural Transfer}

In 2015, The development of the Neural-Style algorithm extends the limit of what a CNN can learn from the picture. By the combination of the content loss and style loss functions, CNN can learn the representation of target image's artistic style. In additional, CNN can reproduce and apply learned image style onto other image. The will help us to modify the image's texture.

\subsection{Transformer}

In 2018, the Google's paper "Attention is All You Need" first introduced attention mechanism and a brand new kind neural layer called Transformer which is completely based on the attention mechanism. And the models which are based on the transformer layer has been proved to be superior in performance in all NLP problems. Other than that, the transformer model's good parallelism resolves the problem that traditional large NLP model is hard to be trained parallel. This breaks the domination of LSTM in NLP applications.

\section{The Designation of Unsupervised Representation Learning Method}

Our aim is to design an unsupervised learning method that we are able to train any large computer vision model on training data without annotations. In Section 3.1, we introduce how to do image representation information sequence self-generating. In Section 3.2, we introduce neural transfer as data augmentation for increasing object shape bias. In Section 3.3, we describe our innovative contrastive prediction coding method. In Section 3.4, the designation of cost function is introduced. In the final Section 3.5, we describe the complete improved contrastive prediction coding architecture and the related algorithm.

\subsection{Image Representation Information Sequences Self-generation} 

\begin{figure}
	\begin{center}
		\includegraphics[width=0.5\textwidth]{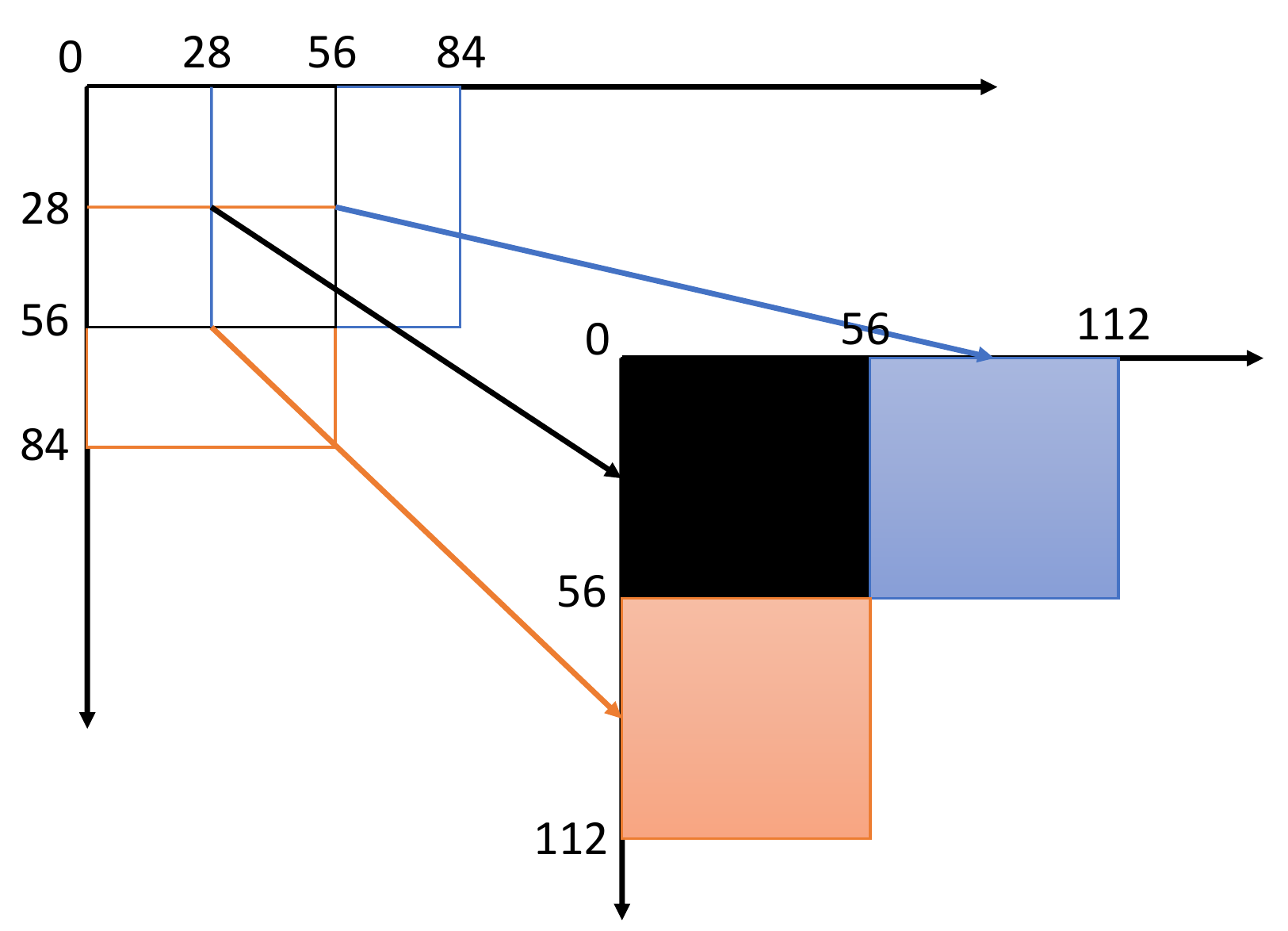}
	\end{center}
	\caption{The generation of image patches grid. For an image after being resized as $224\times224$, the image patches of it are extracted and half overlapped with every patch's neighbor patch. All patches form a $7\times7$ image patches grid eventually.}
	\label{fig:processing}
\end{figure}

\begin{figure}[ht]
	\begin{center}
		\includegraphics[width=0.5\textwidth]{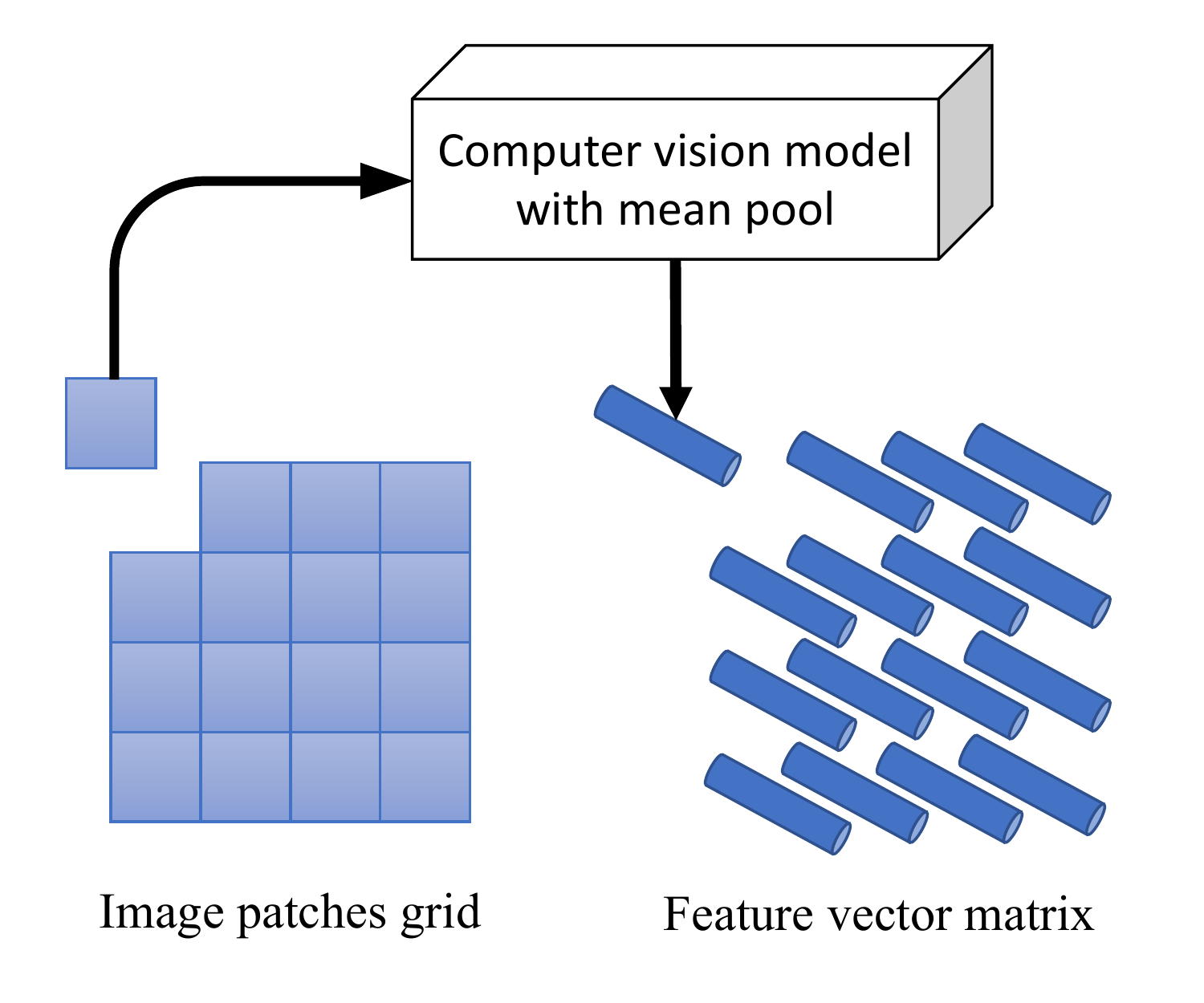}
	\end{center}
	\caption{The generation of image patches grid. Every image patch is overlapped with its neighbor patch as much as its half area. This forms a $7\times7$ image patches grid.}
	\label{fig:processing2}
\end{figure}

As described in {\bf figure \ref{fig:processing}}, every image will first to be resized as shape $224\times224$. Then follow the row and column both directions as shown in {\bf figure \ref{fig:processing}}, with step size as $28$ pixels each time, totally $7\times7$ image  patches with overlap(for the row direction, the overlap area size is $56\times28$, for the column direction, it is $28\times56$) are cropped and formed a new $7\times7$ image patch grid. After that, as shown in {\bf \ref{fig:processing2}}, each element of the image patch grid with size $56\times56$ will be processed by computer vision model's encoder part as image representation learning process, with the mean pool function, this forms a feature vector matrix with size $7\times7\times representation\, infomraiton's\, dimensions$.

\subsection{Data Augmentation}

In order to force computer vision model to learn more complex representation of the object shape. Texture information should be treated as negative information and separated from the shape information. For this purpose, $5$ hand-picked texture samples are chose as target image styles learned by $5$ neural transfer models. By utilizing the $5$ models, $5$ images with different kinds texture can be produced from the original image. All these extra images will be processed in the same way introduced in Section 3.1.

\subsection{Contrastive Prediction Methods}

\begin{figure}
	\begin{center}
		\includegraphics[width=0.5\textwidth]{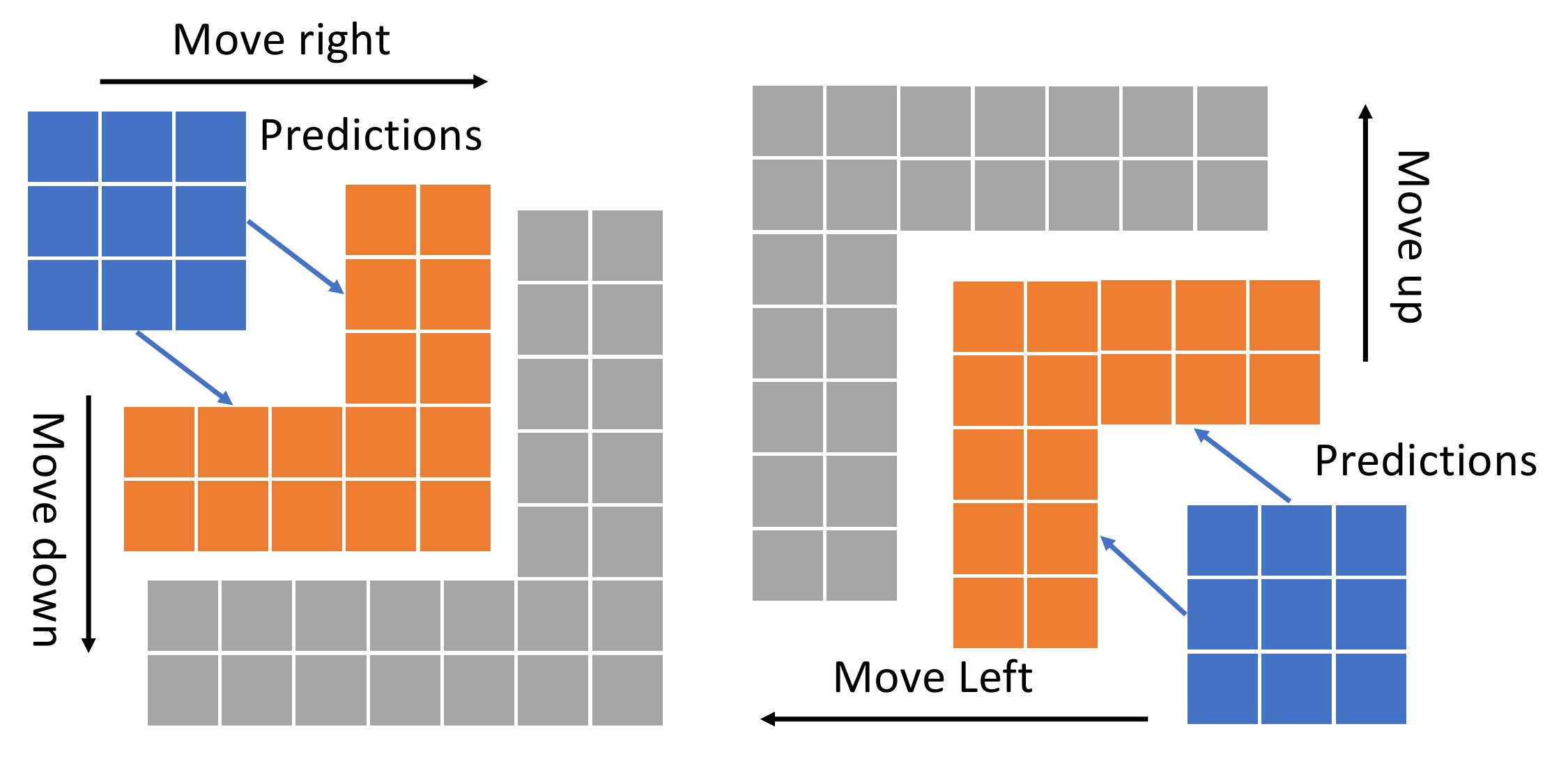}
	\end{center}
	\caption{The illusion of autoregression training on the same image. The blue part of the image patch grid is used as training information to predict the orange part which is around the blue part by the autoregression model. Repeat doing this by moving right or down with one image patch as a step shown as the left example. The right example is the process with opposite direction. }
	\label{fig:training}
\end{figure}

Tradition contrastive prediction method extracts information and make prediction along with the sequence order. However, image information structure is different from context or audio signal whose location information is as same as the sequence order, which comes with strong aggregating attribute. So, in order to make the image patches sequence without losing the image location information, we design the image sequence generation mechanism as show in {\bf figure \ref{fig:training}} and {\bf figure \ref{fig:training2}}: with considering the computing resource requirement, we choose size $3\times3$ image patches as a training block, the two layers of images patches around the training block are the target to predict. Then for each image patch which is chose as the training or target, it will be padded with value $ 0 $ as size $ 224\times224 $, then processed by the computer vision model's encoder part and mean pool layer with output as a vector includes the image patch's representation information. So a sequence of vectors for training is used by the autoregression model which is composed with transformer layers to predict the orange part's vector sequence. This forces model to learn the high level latent representation of the object which can be used to predict object's profile's representation information.   

\subsubsection{Contrastive Learning on Same Image}

An image patches grid  $X$ with size $s\times s$, and $stride=1$, $perception=k$,  $anchor=(i,j)$, the training sequence will be $S_{train}=\{X_{i:i+k,j:j+k}\}$ totally $k^{2}$ image patches from the image grid. The target sequence will be $S_{target}=\{X_{i:i+k+2,j:j+k+2}-X_{i:i+k,j:j+k}\}$ which also meets that $\max (i+k+2,j+k+2)\le s$. With computer vision model $f_{cv}$ and mean pool function, the training sequence will be transferred as a sequence of latent representation vector $\breve{x}=pool(f_{cv}(x))$ for $x\in S_{train}$, so is the target sequence $\breve{y}=pool(f_{cv}(y))$ for $y\in S_{target}$. 

As we have the shape of the sequences, an autoregression function $g_{auto}$ can be constructed. By feeding the model with the transferred training sequence, a prediction sequence $\bar{y}=g_{auto}(S_{train})$ which has the same size of target sequence $\breve{y}$ is produced. By comparing $\bar{y}$ and $\breve{y}$, both models $f_{cv}$ and  $g_{auto}$ are updated during the back propagation process.

\subsubsection{Contrastive Learning On Different Image}
\begin{figure}[h]
	\begin{center}
		\includegraphics[width=0.48\textwidth]{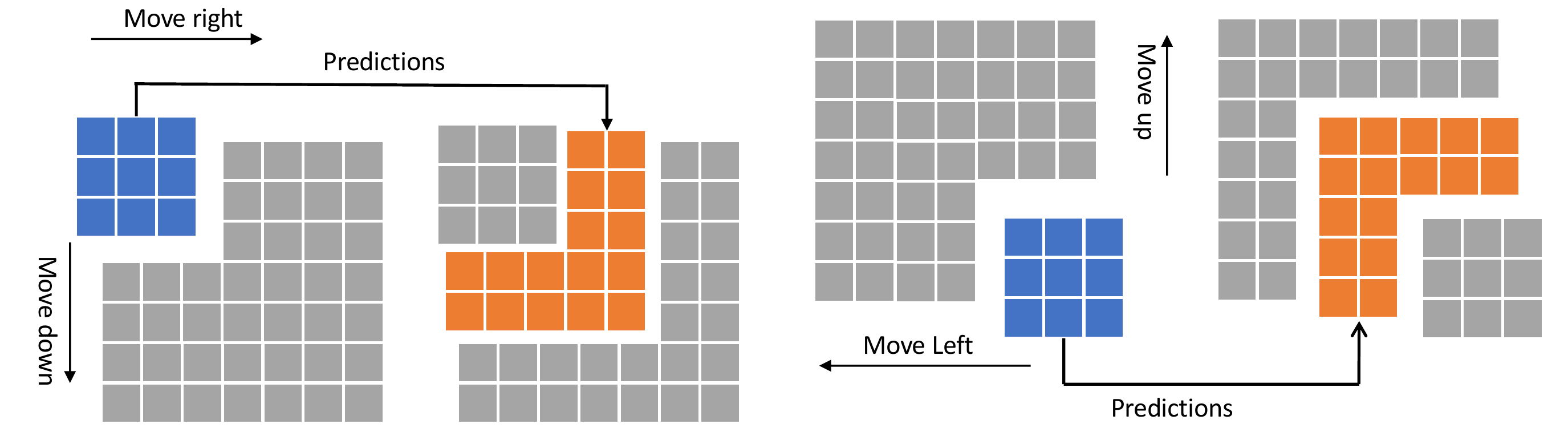}
	\end{center}
	\caption{The illusion of contrastive learning on different images. The blue part which is the training block is always chose from the original image, but the orange part which is the target part is chose from the images which are the original image with different kinds texture. The left and right examples are opposite training directions like the {\bf figure \ref{fig:training}}.}
	\label{fig:training2}
\end{figure}

In order to force model extract feature representation information with more object shape bias, as shown in {\bf figure \ref{fig:training2}}, we utilize the contrastive learning on images with different textures as a regularization method. The process is close in {\bf section 3.3.1} except the target sequence is from different image source. By minimizing the prediction cost of autoregression model, the texture representation information of both training and target sequences should be filtered out during the feature representation information extraction stage with the computer vision model.

\subsection{The Designation of Cost Function}

The training of feature extraction ability of computer vision model is guided by the evaluation of the quality of the prediction of the autoregression model. The essential purpose of the training is a condition probability function that is based on the representation learning information of sampled image patches to predict the target image patches' representation information.  Also, considering the contrastive learning processes on images with different and same texture are running synchronize, the final cost function is a combination of multiple cost functions. The base component of the cost function is based on the regular cross-entropy loss with softmax as the probability prediction function, and it sums the loss over the locations of prediction and target sequences:

 \begin{align}
  L_{c} &=-\log p(\bar{y}|\breve{y},\breve{x})\\
  &=-\sum_{i}\log p(\bar{y}_{i}|\breve{y}_{i},\breve{x})\\
  &=-\sum_{i}\log p(\dfrac{\exp (\bar{y}_{i}^{T}\breve{y}_{i})}{\exp (\bar{y}_{i}^{T}\breve{y}_{i})+\sum_{j}\exp (\breve{y}_{i}\breve{x}_{j})})
 \end{align}

Due to the fact that multiple learning processes happen at the same time on multiple images, the final cost function is composed by multiple cost functions as the follow:
\begin{align}
L_{C}=\omega_{0}L_{c_{0}}+\sum_{i} \omega_{i}L_{c_{i}}
\end{align} 

Where $ L_{c_{0}} $ represents the cost function of contrastive learning on the same image, the $ L_{c_{i}} $ means the cost function of learning on the original image and image with $i$th kind texture. The sum of all $\omega$ values are not necessary to be 1, the ratio between $\omega$ and the sum of $\omega_{i}$ controls model learn more on object's texture's information or shape's.

\subsection{Unsupervised Learning Architecture}

\begin{figure}
	\begin{center}
		\includegraphics[width=0.5\textwidth]{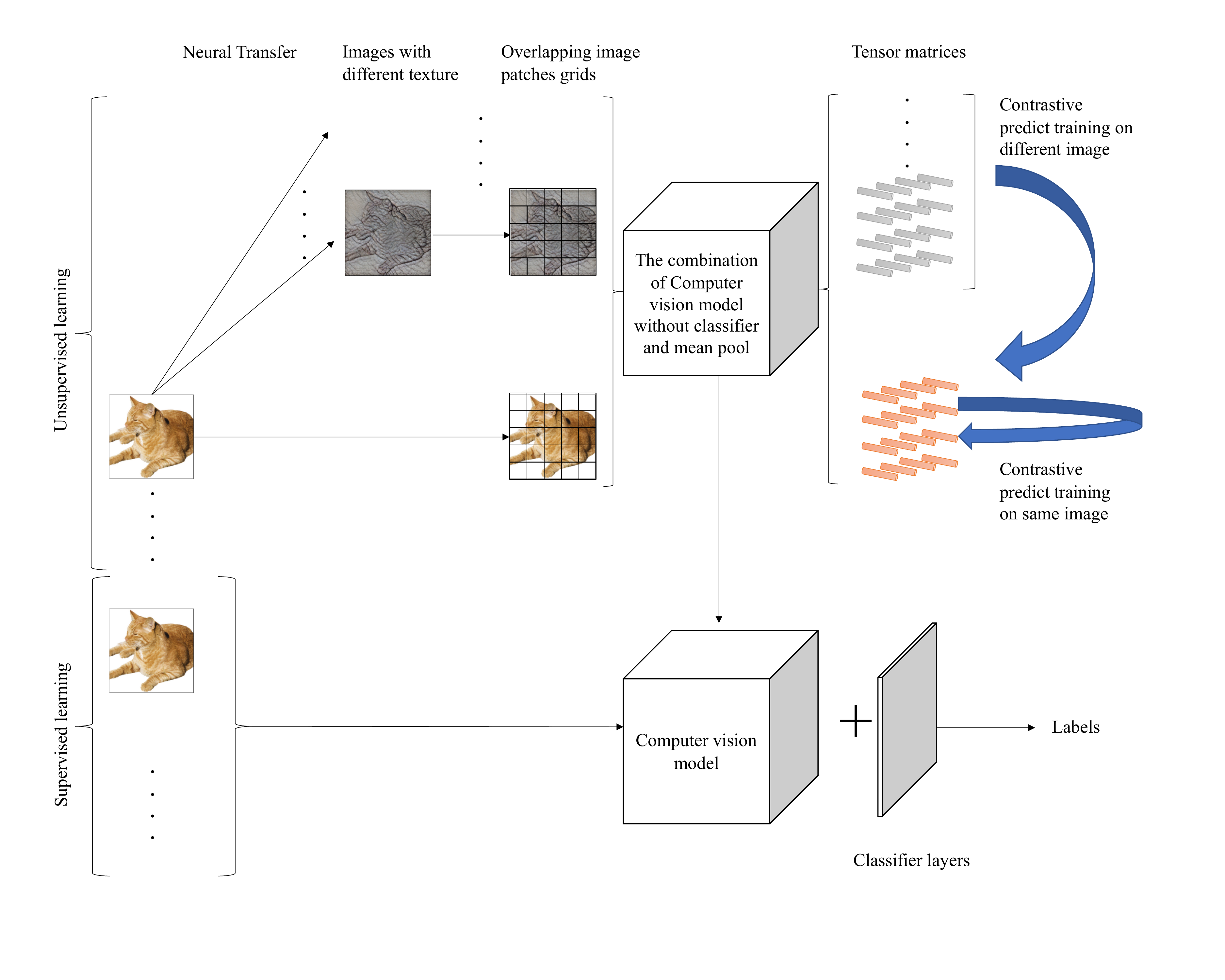}
	\end{center}
	\caption{The designation of final unsupervised training architecture.}
	\label{fig:archi}
\end{figure}

As shown in {\bf Figure \ref{fig:archi}}, for every training image, we used $ 5 $ neural transfer models which are introduced in Section 3.2, to generate another $ 5 $ images with different textures. With the image processing method introduced in section 3.1, multiple representation information vector matrices are generated. During the contrastive learning process described in section 3.3 and section 3.3, the computer vision model will be updated and outputted as a pre-trained model. The related algorithm is {\bf algorithm \ref{alg:1}}.

\begin{algorithm}
	\renewcommand{\algorithmicrequire}{\textbf{Input:}}
	\renewcommand{\algorithmicensure}{\textbf{Output:}}
	\caption{The Unsupervised Learning Architecture}
	\label{alg:1}
	\begin{algorithmic}[1]
	\REQUIRE { 
		The images $ \{x_{i}\} $\\
		Computer vision model $ f_{cv} $\\
		Neural Transfer models $\{f_{nt_j}\}$\\
		Autoregression model $ g_{auto} $\\
		Cost function $F_{unsuper}$}
	\ENSURE Computer vision model $ f_{cv} $
	\STATE {\emph \# Traversing the training images without annotations}
	\STATE {\bf For $ x $ in $ \{x_{i}\} $:}
	\STATE \quad {\emph \# Images with different textures generating}
	\STATE \quad \quad $ x  \xrightarrow{\{f_{nt_j}\}}   \{x^{j}\}$ 
	\STATE \quad  {\emph \# Image patches sequences generating}
	\STATE \quad  $x, \{x^{j}\} \xrightarrow{Sequence} x_{train_{seq}}, x_{target_{seq}}, \{x^{j}_{target_{seq}}\}$
	\STATE \quad \# Representation learning of image patch sequences
	\STATE \quad  $x_{train_{seq}}\xrightarrow{f_{cv}+mean\,pool} v_{train_{seq}}$
	\STATE \quad  $x_{target_{seq}}\xrightarrow{f_{cv}+mean\,pool} v_{target_{seq}}$
	\STATE \quad  $\{x^{j}_{target_{seq}}\}\xrightarrow{f_{cv}+mean\,pool} \{v^{j}_{target_{seq}}\}$
	\STATE \quad  $cost = F_{unsuper}(($\\
	$v_{target_{seq}},\{v^{j}_{target_{seq}}\}), g_{auto}(v_{train_{seq}}))$
	\STATE \quad  Backpropagation of $cost$ to update $f_{cv}, g_{auto}$
	\STATE {\bf End for}
	\STATE {\bf Repeat the FOR loop until the process converges}
	\STATE \textbf{Return} $f_{cv}$
\end{algorithmic}
\end{algorithm}

With appending the classifier layer to the pre-trained computer vision model from previous training, the new model can be used to do supervised learning on any other data set with annotations. The related algorithm is {\bf algorithm \ref{alg:2}}.

\begin{algorithm}
	\renewcommand{\algorithmicrequire}{\textbf{Input:}}
	\renewcommand{\algorithmicensure}{\textbf{Output:}}
	\caption{The Supervised Learning Architecture}
	\label{alg:2}
	\begin{algorithmic}[1]
		\REQUIRE { 
			The images $ \{x_{i}\} $ with annotation\\
			Computer vision model $ f_{cv} $\\
			Cost function $F$}
		\ENSURE Computer vision model $ f_{cv} $
		\STATE {\emph \# Traversing the training images with annotations}
		\STATE {\bf For $ x $ in $ \{x_{i}\} $:}
		\STATE \quad  $cost = F(label, f_{cv}(x))$
		\STATE \quad  Backpropagation to update $f_{cv}$
		\STATE {\bf End for}
		\STATE {\bf Repeat the FOR loop until the process converges}
		\STATE \textbf{Return} $f_{cv}$
	\end{algorithmic}
\end{algorithm}

\section{Experiments}
\subsection{From Imagenet to Imagenet}

\subsection{From Webvision to Webvision}

\subsection{From Webvision to Imagenet}

\subsection{References}

List and number all bibliographical references in 9-point Times,
single-spaced, at the end of your paper. When referenced in the text,
enclose the citation number in square brackets, for
example~\cite{Authors14}.  Where appropriate, include the name(s) of
editors of referenced books.

\begin{table}[h]
\begin{center}
\begin{tabular}{|l|c|}
\hline
Method & Frobnability \\
\hline\hline
Theirs & Frumpy \\
Yours & Frobbly \\
Ours & Makes one's heart Frob\\
\hline
\end{tabular}
\end{center}
\caption{Results.   Ours is better.}
\end{table}



{\small
\bibliographystyle{ieee_fullname}
\bibliography{egbib}
}

\end{document}